\documentclass{article} 
\usepackage[final]{colm2026_conference}

\usepackage{microtype}
\usepackage{hyperref}
\usepackage{url}
\usepackage{booktabs}
\usepackage{times}
\usepackage{latexsym}
\usepackage[T1]{fontenc}
\usepackage[utf8]{inputenc}
\usepackage{inconsolata}
\usepackage{graphicx}
\usepackage{enumitem}
\usepackage{float}
\usepackage{algorithm}
\usepackage{algpseudocode}
\usepackage{tcolorbox}
\usepackage{amsmath}
\usepackage{multirow}

\definecolor{darkblue}{rgb}{0, 0, 0.5}
\hypersetup{colorlinks=true, citecolor=darkblue, linkcolor=darkblue, urlcolor=darkblue}

\title{CricBench: A Multilingual Benchmark for Evaluating LLMs in Cricket Analytics}

\author{
  \textbf{Parth Agarwal\textsuperscript{1}},
  \textbf{Prisha Singhal\textsuperscript{1}},
  \textbf{Navya Kommuri\textsuperscript{1}},
  \textbf{Trizal Garg\textsuperscript{1}},
  \textbf{Dhruv Shah\textsuperscript{1}}, \\
  \textbf{Vaibhav Devraj\textsuperscript{1}},
  \textbf{Jagat Sesh Challa\textsuperscript{1}},
  \textbf{Yash Sinha\textsuperscript{1}},
  \textbf{Murari Mandal\textsuperscript{2}},
  \textbf{Dhruv Kumar\textsuperscript{1}} \\
  \textsuperscript{1}BITS Pilani, Pilani Campus, India \\
  \textsuperscript{2}KIIT Bhubaneswar, India \\
  \\
  \small{\texttt{f20240726@pilani.bits-pilani.ac.in}}
}

\begin{document}

\maketitle
\fancyhead{}

\begin{abstract}
Cricket is the second most popular sport worldwide, with a massive following of billions of fans. Enthusiasts and analysts frequently seek advanced statistical insights such as long-term historical performance trends or complex player comparisons which are often unavailable through standard web searches. Although large language models (LLMs) have advanced significantly in Text-to-SQL tasks, their capability to handle domain-specific nuances, complex schema variations, and multilingual requirements inherent to sports analytics remains under-explored. To investigate this capability gap, we present \textbf{CricBench}, a comprehensive benchmark suite for evaluating the \emph{intrinsic} SQL generation abilities of LLMs on specialized cricket data across four formats: Test, ODI, T20I, and IPL. We curate a Gold-Standard dataset comprising \textbf{2,654 evaluation instances} across four languages (English, Hindi, Punjabi, and Telugu) through collaboration with domain experts in cricket and SQL. We evaluate seven state-of-the-art models including GPT-5 Mini, Claude Sonnet 4, DeepSeek R1 and V3, Qwen 235B, Llama 3.1, and Gemma 2 using a strict evaluation protocol with \emph{schema-only prompting}. Our results reveal that \textbf{no single model dominates across all formats}: GPT-5 Mini leads on Test cricket among full-set models (12.4\% \textbf{Data Match Accuracy}, or \textbf{DMA} -- the fraction of queries returning correct results), while Qwen 235B achieves the highest DMA on IPL (28.7\%) and T20I (17.5\%), and all models achieve 0\% on hard ODI queries. Even the best models exhibit a stark disconnect between syntactic validity ($>$98\% execution accuracy) and semantic correctness ($<$29\% DMA). Comparison against the BIRD benchmark reveals a domain gap of 37-55\% points, demonstrating that current LLMs fundamentally lack the domain reasoning needed for specialized analytics. To the best of our knowledge, CricBench is the first Text-to-SQL benchmark specifically designed for cricket analytics.
\end{abstract}

\section{Introduction}
Large Language Models (LLMs) have achieved significant milestones in translating natural language into structured query language (Text-to-SQL), fundamentally altering how non-technical users interact with databases \citep{codex, gpt4, gpt5mini}. Pioneering benchmarks such as WikiSQL \citep{wikisql} and Spider \citep{spider} have driven rapid improvements in schema linking and syntactic correctness \citep{ratsql}. More recently, the BIRD benchmark \citep{bird} has pushed the frontier by integrating external knowledge reasoning. However, while general-purpose models excel at standard business queries, the extent to which these capabilities transfer to highly specialized domains remains an open research question \citep{domain_gap}.

Sports analytics, particularly cricket, represents a unique testbed for investigating this domain gap. Cricket is a global phenomenon with billions of fans, yet it remains under-represented in computational linguistics research \citep{sjsp_cricket}. Unlike general knowledge questions, this domain necessitates granular reasoning involving precise mathematical operations, temporal awareness, compound derived metrics, and phase-specific filters. Furthermore, the domain presents significant entity resolution challenges, such as resolving dynamic player affiliations across varying temporal contexts and mapping players to specific national teams in different eras, thereby testing an LLM's ability to maintain knowledge consistency \citep{cot}.

Current Text-to-SQL benchmarks are predominantly English-centric \citep{spider, bird}. This limitation is particularly critical for sports like cricket, where the fanbase is linguistically diverse. In regions like India, users often interact in native languages or code-mixed dialects. A model that generates accurate SQL for an English query but fails to process the same intent in Hindi is of limited utility in real-world deployment \citep{multilingual_sql, indic_sql}.

To address these challenges, we introduce \textbf{CricBench}, a specialized, multilingual benchmark designed to evaluate the \emph{intrinsic} reasoning capabilities of LLMs in cricket analytics. Rather than augmenting models with domain-specific prompts or retrieval systems, CricBench tests models with \textbf{schema-only prompting}, providing only the database schema and minimal instructions to measure their raw ability to translate domain-specific natural language into correct SQL. CricBench spans \textbf{four cricket formats}: three international formats (Test, ODI, and T20I), where teams represent nations and nomenclature is stable, as well as the Indian Premier League (IPL), a domestic franchise-based T20 tournament with distinct schema characteristics. In total, CricBench comprises \textbf{2,654 evaluation instances} across four languages: English, Hindi, Punjabi, and Telugu. We evaluate a suite of frontier models, including GPT-5 Mini \citep{gpt5mini}, Claude Sonnet 4 \citep{claude}, DeepSeek R1 \citep{deepseekr1} and V3 \citep{deepseekv3}, Llama 3.1 \citep{llama3}, Gemma 2 \citep{gemma}, and Qwen 235B \citep{qwen}, to provide a comprehensive assessment of the current state of AI in sports intelligence.

Our primary contributions are as follows:
\begin{itemize}[noitemsep]
    \item \textbf{CricBench Dataset:} We release the first expert-curated, domain-specific Text-to-SQL benchmark for cricket analytics, comprising 2,654 evaluation instances across four cricket formats with labeled difficulty levels.\footnote{The dataset and code will be made publicly available upon acceptance.}
    \item \textbf{Multilingual Support:} We introduce verified test sets in four languages: English, Hindi, Punjabi, and Telugu which incorporate authentic code-mixing, addressing the lack of Indic language resources in specialized SQL benchmarks.
    \item \textbf{Cross-Format Analysis:} We reveal that model performance varies dramatically across cricket formats. No single model dominates all four, suggesting that format-specific characteristics pose distinct reasoning challenges.
    \item \textbf{Intrinsic Capability Assessment:} By using schema-only prompting without domain-specific context injection, we isolate and measure the inherent reasoning capabilities of LLMs, revealing that even frontier models achieve below 29\% DMA on specialized cricket queries.
    \item \textbf{Domain Gap Quantification:} We benchmark the same models against BIRD \citep{bird} and demonstrate a massive domain gap. Models lose 37-55\% points of accuracy when moving from general-purpose to cricket-specific queries (Section~\ref{sec:domain_gap}).
    \item \textbf{Expert Curation:} Unlike recent trends in synthetic data generation, our Gold Standard is manually authored and verified by domain experts to ensure logical and factual correctness.
\end{itemize}

\section{Related Work}
The development of Natural Language Interfaces to Databases (NLIDB) has been a longstanding goal in NLP. Our work builds upon and extends research in three key areas: Text-to-SQL generation, domain-specific benchmarking, and multilingual evaluation.

\subsection{Text-to-SQL Benchmarks}
The standard for evaluating Text-to-SQL systems has historically been set by large-scale, cross-domain datasets. WikiSQL \citep{wikisql} provided a foundation for simple SELECT-based queries, while Spider \citep{spider} introduced complex, multi-table joins and nested queries, becoming the de-facto benchmark for measuring generalization across unseen domains. More recently, BIRD \citep{bird} has pushed the frontier by focusing on dirty database contents and external knowledge reasoning. \citet{lei2025texttosql} provide a comprehensive survey of Text-to-SQL methods including domain-specific challenges. While these benchmarks are invaluable for general-purpose evaluation, they lack the depth required for intense domain specificity found in specialized fields like sports analytics.

\subsection{Domain-Specific Query Generation}
Research has increasingly demonstrated that generalist models often struggle with the nuances of specialized domains like healthcare and finance, where correct SQL generation depends heavily on understanding domain logic rather than just schema linking \citep{domain_gap}. Previous works have attempted to mitigate this via Retrieval-Augmented Generation (RAG) \citep{rag} and few-shot prompting \citep{fewshot}. In the sports domain, \citet{sjsp_cricket} assessed the accuracy of LLMs in extracting cricket information, finding that models struggle with precise statistical retrieval. However, that work focuses on information extraction from unstructured text rather than structured query generation against relational databases. \citet{naacl2024_sql} have investigated domain adaptation for structured query generation. \textbf{To the best of our knowledge, no prior work has proposed a Text-to-SQL benchmark specifically targeting cricket analytics.} Our work fills this gap with a rigorous, expert-curated benchmark, deliberately using \textbf{minimal prompting} rather than engineered prompts to expose the inherent limitations of current LLMs when faced with domain-specific reasoning challenges.

\subsection{Multilingual Evaluation}
While English-centric Text-to-SQL research is mature, multilingual benchmarks remain scarce for Indic languages; existing resources often focus on high-resource languages or rely on machine-translated versions of general benchmarks \citep{multilingual_sql}. Recent efforts such as Indic Text-to-SQL datasets \citep{indic_sql} have begun to address this gap, but coverage of domain-specific scenarios remains limited. Our work extends this line of research by introducing verified English, Hindi, Punjabi, and Telugu test sets for a specialized domain. We specifically address the challenge of code-mixing, where technical terms are retained in English (e.g., ``Strike Rate'') within Indic sentence structures, mirroring authentic user behavior in the Indian subcontinent.

\section{Methodology}
\textbf{CricBench} is a benchmark suite encompassing data engineering, expert curation, and a standardized evaluation pipeline. The system architecture is designed to test not just SQL generation, but also domain-specific reasoning and multilingual understanding.

\subsection{Dataset Construction}
The foundation of CricBench is a set of robust, normalized relational databases covering four cricket formats. We ingested comprehensive ball-by-ball records sourced from Cricsheet \citep{cricsheet}, transforming semi-structured JSON data into strict SQLite schemas \citep{sqlite}. As shown in Table~\ref{tab:dataset_stats}, the dataset comprises 633 unique base queries distributed across Test (169), ODI (64), T20I (200), and IPL (200) formats. Each query is translated into four languages (English, Hindi, Punjabi, and Telugu), yielding a total of 2,654 evaluation instances. The IPL format additionally includes extended Hindi and Telugu translation variants that incorporate different code-mixing patterns, contributing to its higher instance count of 922.

\begin{table}[t]
\begin{center}
\begin{tabular}{lccccc}
\toprule
\textbf{Format} & \textbf{Queries} & \textbf{Easy} & \textbf{Medium} & \textbf{Hard} & \textbf{Instances} \\
\midrule
Test  & 169 & 44 & 79 & 46 & 676 \\
ODI   & 64  & 35 & 24 & 5  & 256 \\
T20I  & 200 & 24 & 92 & 84 & 799 \\
IPL   & 200 & 48 & 112 & 70 & 922 \\
\midrule
\textbf{Total} & \textbf{633} & 151 & 307 & 205 & \textbf{2,654} \\
\bottomrule
\end{tabular}
\end{center}
\caption{CricBench dataset statistics. Each base query exists in English, Hindi, Punjabi, and Telugu. IPL includes additional Hindi and Telugu translation variants with different code-mixing patterns, yielding 922 instances from 200 base queries.}\label{tab:dataset_stats}
\end{table}

The international formats (Test, ODI, T20I) share a common schema consisting of five core tables:
\begin{itemize}[noitemsep]
    \item \textbf{Matches:} Metadata for each game, including dates, venues, and results. Chronology is strictly enforced via \texttt{start\_date}.
    \item \textbf{Deliveries:} Granular ball-by-ball events, serving as the source of truth for all derived statistics.
    \item \textbf{Players:} A central registry for entity resolution (\texttt{player\_id}, \texttt{player\_name}).
    \item \textbf{PlayerInMatch:} A linking table associating players with specific matches and teams.
    \item \textbf{FielderDismissals:} A specialized junction table linking dismissal events to specific fielders.
\end{itemize}

The IPL database uses a similar but distinct schema tailored to franchise-based cricket, where teams represent franchises rather than nations and season structures differ from international formats.

\subsection{Gold Standard Curation}
To ensure the benchmark's reliability, we eschewed automated generation in favor of a rigorous \textbf{manual curation process}.

\subsubsection{Expert Authoring \& Verification}
We collaborated with domain experts in cricket analytics who have represented their universities in the sport and participated in inter-university tournaments across India, to manually author high-complexity natural language questions derived from real-world analysis patterns observed in match commentaries and statistical leaderboards on ESPNcricinfo \citep{espn} and Cricbuzz \citep{cricbuzz}.

\subsubsection{Difficulty Taxonomy \& Distribution}
We assign each query an overall difficulty label (\textit{Easy}, \textit{Medium}, \textit{Hard}) based on the SQL features and domain knowledge it requires. Table~\ref{tab:complexity} summarizes the prevalence of individual complexity dimensions across the dataset.

\begin{table}[t]
\begin{center}
\begin{tabular}{lcc}
\toprule
\textbf{Complexity Category} & \textbf{Count} & \textbf{\% of Dataset} \\
\midrule
Aggregations (Group By/Having) & 379 & 59.9\% \\
Multi-table Joins              & 358 & 56.6\% \\
Temporal Filtering             & 238 & 37.6\% \\
Nested Queries / CTEs          & 194 & 30.6\% \\
Derived Metrics (SR, Econ)     & 114 & 18.0\% \\
Window Functions (Rank etc.)   & 32  & 5.1\%  \\
\bottomrule
\end{tabular}
\end{center}
\caption{Distribution of query complexity across all 633 CricBench base queries. Categories are not mutually exclusive; a single query often involves joins, temporal logic, and derived metric computation.}\label{tab:complexity}
\end{table}

\textbf{Verification:} For every question, the corresponding SQL query was hand-written. The answers were manually cross-referenced against authentic third-party records from Cricbuzz \citep{cricbuzz} and ESPNcricinfo \citep{espn} to guarantee factual correctness.

\textbf{Multilingual Expansion:} The dataset was manually translated into Hindi, Punjabi, and Telugu, specifically preserving technical terms via code-mixing (e.g., retaining ``Strike Rate'' in English script) to mirror authentic user behavior.

An illustrative example from the Test cricket Easy category is: \emph{``Who scored the most runs in Test matches played at Lord's?''} The corresponding SQL requires joining three tables with a venue filter, GROUP BY, and ORDER BY with LIMIT; simple in structure but requiring correct schema navigation. See Appendix~\ref{sec:appendix_examples} for examples across all formats, difficulty levels, and languages.

\subsection{Evaluation Protocol}
\label{sec:eval_protocol}
To measure the \emph{intrinsic} SQL generation capability of each model, we employ a \textbf{schema-only prompting} strategy. Each model receives an identical system prompt containing the database schema definition and minimal instructions (see Appendix~\ref{sec:appendix_prompts}). The prompt includes only two lightweight domain hints -- the definition of a legal delivery (\texttt{wides=0 AND noballs=0}) and the exclusion list for bowler wickets, which disambiguate schema semantics but do \emph{not} provide calculation formulas, statistical definitions, or few-shot examples. This design choice is deliberate: by withholding substantive domain context, we isolate each model's inherent ability to reason about specialized queries from its ability to follow explicit instructions.

\textbf{Full-Set vs.\ Subset Evaluation.} Due to computational and API cost constraints, not all models were evaluated on the complete benchmark. We define two evaluation configurations:
\begin{itemize}[noitemsep]
    \item \textbf{CricBench-Full:} Models evaluated on the complete query set for all (or most) formats. This group includes GPT-5 Mini, DeepSeek V3, Gemma 2, Claude Sonnet 4, and Llama 3.1.
    \item \textbf{CricBench-Sub:} Models evaluated on representative \emph{stratified subsets}, sampled to maintain proportional representation across difficulty levels and languages. This group includes DeepSeek R1 and Qwen 235B.
\end{itemize}
Table~\ref{tab:main_results} reports the sample size $N$ for each model--format combination to ensure transparency. All subset selections were designed to preserve the difficulty and language distribution of the full benchmark, ensuring unbiased evaluation despite reduced sample sizes. A small number of CricBench-Full models have reduced sample sizes for specific formats due to API limitations (e.g., Llama 3.1 on ODI with $N$=144, Claude Sonnet 4 on IPL with $N$=130).

\subsection{Evaluation Metrics}
We evaluate model performance using two primary metrics:

\begin{itemize}[noitemsep]
    \item \textbf{Execution Accuracy (EX):} The percentage of generated SQL queries that execute against the database without syntax or runtime errors.

    \item \textbf{Data Match Accuracy (DMA):} The percentage of generated SQL queries whose results \emph{exactly match} the Gold Standard output. This strict metric serves as the primary indicator of semantic correctness.
\end{itemize}

\section{Results and Discussion}

\subsection{Cross-Format Overview}
Table~\ref{tab:main_results} presents the main results across all four cricket formats. Several striking patterns emerge.

\begin{table}[t]
\begin{center}
\scriptsize
\setlength{\tabcolsep}{2.5pt}
\begin{tabular}{l|ccc|ccc|ccc|ccc}
\toprule
& \multicolumn{3}{c|}{\textbf{Test}} & \multicolumn{3}{c|}{\textbf{ODI}} & \multicolumn{3}{c|}{\textbf{T20I}} & \multicolumn{3}{c}{\textbf{IPL}} \\
\textbf{Model} & $N$ & \textbf{EX} & \textbf{DMA} & $N$ & \textbf{EX} & \textbf{DMA} & $N$ & \textbf{EX} & \textbf{DMA} & $N$ & \textbf{EX} & \textbf{DMA} \\
\midrule
\multicolumn{13}{l}{\textit{CricBench-Full}} \\
GPT-5 Mini      & 676 & 99.7 & 12.4 & 256 & 98.8 & 10.9 & 799 & 98.9 & 2.8  & 922 & 97.5 & 11.0 \\
DeepSeek V3     & 676 & 99.9 & 11.8 & 256 & 95.7 & 10.5 & 799 & 85.7 & 7.9  & 922 & 95.2 & 8.0 \\
Claude Sonnet 4 & 676 & 92.5 & 8.9  & 256 & 93.8 & 6.2  & 799 & 92.7 & 8.4  & 130$^\dagger$ & 83.1 & 23.8 \\
Llama 3.1 (8B)  & 676 & 95.3 & 3.7  & 144$^\dagger$ & 39.6 & 1.4  & 799 & 19.1 & 0.4  & 922 & 56.2 & 0.5 \\
Gemma 2 (9B)    & 676 & 67.8 & 2.4  & 256 & 53.1 & 3.5  & 799 & 56.1 & 0.9  & 922 & 34.3 & 0.3 \\
\midrule
\multicolumn{13}{l}{\textit{CricBench-Sub (stratified subsets)}} \\
DeepSeek R1     & 100 & 98.0 & \textbf{33.0} & 160 & 95.6 & \textbf{11.9} & 199 & 96.5 & 4.5  & 230 & 95.7 & 12.6 \\
Qwen 235B        & 168 & 98.8 & 10.1 & 160 & 27.5 & 5.6  & 200 & 79.5 & \textbf{17.5} & 230 & 88.7 & \textbf{28.7} \\
\bottomrule
\end{tabular}
\end{center}
\caption{Cross-format performance with schema-only prompting. $N$ = evaluation instances. EX = Execution Accuracy (\%), DMA = Data Match Accuracy (\%). Bold DMA values indicate the best-performing model per format. $\dagger$ indicates a reduced sample due to API limitations. CricBench-Sub models were evaluated on stratified subsets; direct comparison with CricBench-Full requires caution.}\label{tab:main_results}
\end{table}

\paragraph{No Single Model Dominates.}
The most important finding is that \textbf{model rankings shift substantially across formats}. Among CricBench-Full models, GPT-5 Mini leads on Test (12.4\%) and ODI (10.9\%), DeepSeek V3 performs competitively on T20I (7.9\%) and IPL (8.0\%), and Claude Sonnet 4 achieves 23.8\% on its IPL subset ($N$=130). Among CricBench-Sub models, DeepSeek R1 achieves a striking 33.0\% on its Test subset ($N$=100) but only 4.5\% on T20I, while Qwen 235B leads T20I (17.5\%) and IPL (28.7\%) but struggles on ODI (5.6\% DMA, 27.5\% EX). This format-dependent variation suggests that different cricket formats pose qualitatively different reasoning challenges that engage different model capabilities.

\paragraph{The Execution-Accuracy Disconnect.}
A pervasive finding is the vast gap between execution accuracy and data match accuracy. GPT-5 Mini and DeepSeek V3 achieve near-perfect execution rates ($>$99\% on Test) yet return correct results only $\sim$12\% of the time. This ``illusion of competence'' demonstrates that \textbf{syntactic validity is a poor proxy for semantic correctness}: models readily produce well-formed queries that reference correct table and column names but fail to implement the domain logic required for correct results.

\paragraph{Small Models Degrade Severely.}
Both Llama 3.1 (8B) and Gemma 2 (9B) exhibit severe degradation on T20I and IPL. Llama 3.1 drops from 95.3\% EX on Test to 19.1\% on T20I, and Gemma 2 falls to 34.3\% EX on IPL. These models also achieve near-zero DMA on T20I and IPL ($\leq$0.9\%), suggesting that the more complex schemas and query types in these formats exceed the capacity of smaller models.

\paragraph{IPL as a Differentiator.}
The IPL format reveals performance patterns distinct from international formats. Qwen 235B and Claude Sonnet 4 achieve their highest DMA on IPL (28.7\% and 23.8\% respectively), significantly above their performance on other formats. GPT-5 Mini maintains consistent performance (11.0\%), while small models collapse. The IPL schema's franchise-based structure and distinct match metadata appear to exercise different model capabilities than international cricket.

\subsection{Difficulty Analysis}
\label{sec:difficulty}
Table~\ref{tab:difficulty} presents the DMA breakdown by difficulty level across formats.

\begin{table}[t]
\begin{center}
\scriptsize
\setlength{\tabcolsep}{2pt}
\begin{tabular}{l|ccc|ccc|ccc|ccc}
\toprule
& \multicolumn{3}{c|}{\textbf{Test}} & \multicolumn{3}{c|}{\textbf{ODI}} & \multicolumn{3}{c|}{\textbf{T20I}} & \multicolumn{3}{c}{\textbf{IPL}} \\
\textbf{Model} & \textbf{E} & \textbf{M} & \textbf{H} & \textbf{E} & \textbf{M} & \textbf{H} & \textbf{E} & \textbf{M} & \textbf{H} & \textbf{E} & \textbf{M} & \textbf{H} \\
\midrule
GPT-5 Mini      & 13.6 & 11.4 & 13.0 & 13.6 & 9.4  & 0.0 & 0.0  & 8.6  & 1.4 & 16.7 & 8.9  & 10.3 \\
DeepSeek V3     & 11.9 & 11.1 & 13.0 & 15.7 & 5.2  & 0.0 & 1.0  & 10.3 & 7.2 & 13.5 & 7.3  & 5.3 \\
Claude Son.\ 4  & 6.2  & 12.7 & 4.9  & 6.4  & 7.3  & 0.0 & 6.3  & 12.2 & 4.8 & 31.7 & 27.3 & 13.3 \\
Llama 3.1       & 12.5 & 0.9  & 0.0  & 4.2  & 0.0  & 0.0 & 0.0  & 0.3  & 0.6 & 2.6  & 0.0  & 0.0 \\
Gemma 2         & 5.1  & 1.9  & 0.5  & 5.0  & 2.1  & 0.0 & 0.0  & 0.8  & 1.2 & 1.6  & 0.0  & 0.0 \\
\midrule
DeepSeek R1     & 25.0 & 50.0 & 25.0 & 17.9 & 10.7 & 0.0 & 3.1  & 5.9  & 4.5 & 34.4 & 2.7  & 8.5 \\
Qwen 235B        & 10.7 & 9.4  & 11.4 & 14.3 & 1.2  & 0.0 & 31.2 & 31.8 & 2.1 & 29.5 & 27.3 & 30.5 \\
\bottomrule
\end{tabular}
\end{center}
\caption{DMA (\%) by difficulty level across formats. E=Easy, M=Medium, H=Hard. All models score 0\% on hard ODI queries. GPT-5 Mini T20I difficulty values are from a 399-instance subset. CricBench-Sub models (below midrule) were evaluated on stratified subsets.}\label{tab:difficulty}
\end{table}

\paragraph{Hard ODI Queries Defeat All Models.}
Remarkably, \textbf{every model scores 0\%} on hard ODI queries (though the sample size is small, $N$=20). Even models that perform well on hard queries in other formats completely fail here, suggesting that hard ODI queries require a combination of domain knowledge that no current model possesses.

\paragraph{Inconsistent Performance Across Difficulty Levels.}
Model performance across difficulty levels is generally inconsistent rather than following a predictable pattern. DeepSeek R1 achieves 50.0\% on medium Test queries but only 25.0\% on easy and hard, while on IPL it scores 34.4\% on easy but drops to 2.7\% on medium. Qwen 235B shows strong easy/medium T20I performance (31.2\%/31.8\%) but collapses on hard T20I (2.1\%), while maintaining 30.5\% on hard IPL queries. These patterns suggest that model capabilities interact unpredictably with query characteristics across formats, and that high performance on one difficulty level does not guarantee similar performance on others.

\paragraph{GPT-5 Mini and DeepSeek V3: Consistent but Bounded.}
Both GPT-5 Mini and DeepSeek V3 exhibit relatively \emph{flat} performance across difficulty levels on Test (11-13\% across all levels), suggesting their failures are driven by systematic domain knowledge gaps rather than reasoning complexity. On T20I, DeepSeek V3 shows a notable pattern: 1.0\% on easy, 10.3\% on medium, and 7.2\% on hard, indicating that its medium/hard performance benefits from certain query structures that happen to align with its training.

\subsection{Multilingual Analysis}
\label{sec:multilingual}
Table~\ref{tab:language} presents the language breakdown for Test cricket, where we have the most balanced evaluation data.

\begin{table}[t]
\begin{center}
\small
\begin{tabular}{l|cc|cc|cc|cc}
\toprule
& \multicolumn{2}{c|}{\textbf{English}} & \multicolumn{2}{c|}{\textbf{Hindi}} & \multicolumn{2}{c|}{\textbf{Punjabi}} & \multicolumn{2}{c}{\textbf{Telugu}} \\
\textbf{Model} & EX & DMA & EX & DMA & EX & DMA & EX & DMA \\
\midrule
GPT-5 Mini      & 99.4 & 13.0 & 99.4 & 14.8 & 100  & 11.2 & 100  & 10.7 \\
DeepSeek V3     & 99.4 & 14.2 & 100  & 13.0 & 100  & 11.8 & 100  & 8.3 \\
DeepSeek R1     & 96.0 & 32.0 & 100  & 24.0 & 100  & 36.0 & 96.0 & 40.0 \\
Qwen 235B        & 97.6 & 7.1  & 100  & 11.9 & 97.6 & 11.9 & 100  & 9.5 \\
Claude Sonnet 4 & 92.9 & 8.3  & 91.7 & 9.5  & 94.1 & 8.9  & 91.1 & 8.9 \\
Llama 3.1       & 95.3 & 5.3  & 95.9 & 4.1  & 94.7 & 4.1  & 95.3 & 1.2 \\
Gemma 2         & 57.4 & 3.6  & 62.7 & 1.8  & 66.3 & 1.8  & 84.6 & 2.4 \\
\bottomrule
\end{tabular}
\end{center}
\caption{Performance by language on Test cricket. All values are percentages. DeepSeek R1 is evaluated on a 100-instance subset ($N$=25 per language).}\label{tab:language}
\end{table}

\paragraph{Uniformly Low Performance Across Languages.}
The most salient observation is that \textbf{all models perform poorly across all four languages}. With DMA values below 15\% for CricBench-Full models, the absolute performance gaps between languages are small and unlikely to be statistically significant given the sample sizes (typically 25-200 instances per cell). While some models show slightly higher DMA in one language versus another (e.g., GPT-5 Mini: 14.8\% Hindi vs.\ 10.7\% Telugu on Test), these differences amount to only a few additional correct queries out of 169 and should not be over-interpreted.

\paragraph{Code-Mixed Queries Do Not Degrade Performance.}
Multilingual code-mixed queries do not show systematic degradation relative to pure English, indicating that models handle code-mixed input comparably well. However, the low DMA across all languages suggests that the main limitation is domain-specific reasoning rather than language understanding.

\subsection{Model-Level Analysis}

Table~\ref{tab:model_comparison} summarizes the model taxonomy alongside aggregate performance.

\begin{table}[t]
\begin{center}
\small
\begin{tabular}{lcccc}
\toprule
\textbf{Model} & \textbf{Params} & \textbf{Type} & \textbf{Total $N$} & \textbf{Agg DMA} \\
\midrule
\multicolumn{5}{l}{\textit{Proprietary / Frontier}} \\
GPT-5 Mini      & Undisclosed & Instr.-tuned  & 2,653 & 8.9\% \\
Claude Sonnet 4 & Undisclosed & Instr.-tuned  & 1,861 & 9.3\% \\
\midrule
\multicolumn{5}{l}{\textit{Open-Weight / Large}} \\
DeepSeek R1     & 671B (MoE)  & Reasoning     & 689   & 13.1\% \\
DeepSeek V3     & 671B (MoE)  & Instr.-tuned  & 2,653 & 9.2\% \\
Qwen 235B        & MoE         & Instr.-tuned  & 758   & 16.8\% \\
\midrule
\multicolumn{5}{l}{\textit{Open-Weight / Small}} \\
Llama 3.1       & 8B          & Instr.-tuned  & 2,541 & 1.4\% \\
Gemma 2         & 9B          & Instr.-tuned  & 2,653 & 1.3\% \\
\bottomrule
\end{tabular}
\end{center}
\caption{Model taxonomy and aggregate performance. Agg DMA is computed across all evaluated instances (varying $N$). DeepSeek R1 and Qwen 235B were evaluated on stratified subsets (CricBench-Sub), which limits direct aggregate comparability.}\label{tab:model_comparison}
\end{table}

\paragraph{Scale vs.\ Architecture.}
Proprietary frontier models (GPT-5 Mini, Claude Sonnet 4) do not uniformly outperform open-weight alternatives. On T20I, DeepSeek V3 (7.9\%) and Claude Sonnet 4 (8.4\%) outperform GPT-5 Mini (2.8\%). On IPL, Claude Sonnet 4 achieves 23.8\% on its subset -- the second-highest DMA on any format among CricBench-Full models. This suggests that for domain-specific SQL generation, model architecture and training approach may matter more than raw scale.

\paragraph{Reasoning vs.\ Instruction-Tuning.}
DeepSeek R1 (reasoning-tuned) and DeepSeek V3 (instruction-tuned) share the same 671B MoE architecture. R1's performance is notably inconsistent across formats: 33.0\% on its Test subset but only 4.5\% on T20I, whereas V3 shows more stable but lower performance (7.9-11.8\% across formats). This inconsistency suggests that the reasoning approach does not uniformly benefit all query types; rather, its effectiveness depends on the specific format and query characteristics.

\subsection{The Domain Gap: General vs.\ Specialized}
\label{sec:domain_gap}
To contextualize CricBench performance, we compare our models against the BIRD development set ($N$=1,534), a leading general-purpose Text-to-SQL benchmark that evaluates execution accuracy (i.e., whether the query result matches the gold standard) \citep{bird}. We adopt BIRD execution accuracy scores from recent vanilla (greedy decoding, no retrieval or agent pipelines) evaluations \citep{databricks_rlvr, arctic_sql}. Since BIRD execution accuracy and CricBench DMA both measure whether a query produces the correct result set, they are directly comparable metrics. Table~\ref{tab:domain_gap} presents the results.

\begin{table}[t]
\begin{center}
\small
\begin{tabular}{lccc}
\toprule
\textbf{Model} & \textbf{BIRD (Gen.)} & \textbf{CricBench (Spec.)} & \textbf{Domain Gap} \\
\midrule
GPT-5 Mini$^*$      & 63.8\% & 8.9\%  & $-$54.9\% \\
Qwen 2.5 72B Instruct (closest available proxy)$^\dagger$  & 60.3\% & 16.8\% & $-$43.5\% \\
DeepSeek R1         & 58.0\% & 13.1\% & $-$44.9\% \\
Claude Sonnet 4$^*$ & 59.3\% & 9.3\%  & $-$50.0\% \\
DeepSeek V3         & 55.6\% & 9.2\%  & $-$46.4\% \\
Llama 3.1 (8B)      & 41.9\% & 1.4\%  & $-$40.5\% \\
Gemma 2 (9B)$^\ddagger$ & 38.3\% & 1.3\%  & $-$37.0\% \\
\bottomrule
\end{tabular}
\end{center}
\caption{The Domain Gap: comparison of accuracy on BIRD dev set \citep{bird} vs.\ CricBench aggregate DMA (schema-only prompting). $^*$BIRD score from closest available model version (GPT-4o, Claude 3.5/3.7 Sonnet). $^\dagger$BIRD score from Qwen 2.5 Coder 32B. $^\ddagger$BIRD score estimated from comparable Gemma-family models.}\label{tab:domain_gap}
\end{table}

\paragraph{Massive and Universal Degradation.}
The most striking finding is that \textbf{every model suffers a dramatic performance drop} when moving from general-purpose to specialized cricket queries. The domain gap ranges from $-$37.0\% (Gemma 2) to $-$54.9\% (GPT-5 Mini), with all models losing at least 37 percentage points of accuracy. This demonstrates that the challenges posed by CricBench -- domain-specific metric computation, entity resolution, and temporal reasoning are fundamentally different from, and substantially harder than, the general schema-linking tasks captured by BIRD.

\paragraph{Larger Models Suffer Larger Absolute Drops.}
Counterintuitively, models with the highest BIRD scores experience the largest absolute degradation. GPT-5 Mini drops 54.9 points (63.8\% to 8.9\%) and Qwen 235B drops 43.5 points (60.3\% to 16.8\%), while Llama 3.1 and Gemma 2 lose 40.5 and 37.0 points respectively, smaller absolute drops from much lower baselines. This suggests that general-purpose Text-to-SQL proficiency does not transfer proportionally to specialized domains: the skills driving BIRD performance (schema linking, standard aggregation patterns) are necessary but far from sufficient for cricket analytics.

\paragraph{No Model Retains Even Half Its General Capability.}
On BIRD, all seven models achieve at least 38\% accuracy, and the best reach $\sim$65\%. On CricBench, the best aggregate DMA is 16.8\% (Qwen 235B, on a stratified subset). No model retains more than 26\% of its BIRD-level accuracy on CricBench, underscoring that \textbf{domain-specific reasoning represents a qualitatively different challenge} that current LLMs cannot address through scale or instruction-tuning alone.

\section{Error Analysis}
To understand the root causes of low DMA despite high execution rates, we conducted a qualitative analysis of failure patterns.

\paragraph{Schema Hallucination.}
Despite being provided with the exact schema, models frequently reference non-existent columns. GPT-5 Mini hallucinates columns such as \texttt{total\_runs} or \texttt{match\_year} that do not exist in the schema. This indicates that models' parametric knowledge of ``typical'' database schemas overrides the explicit schema provided in the prompt.

\paragraph{Derived Metric Errors.}
The most prevalent failure category involves incorrect computation of domain-specific metrics. For instance, calculating \textit{Economy Rate} requires: (a) filtering for legal deliveries (\texttt{wides=0 AND noballs=0}), (b) computing runs conceded, (c) dividing by legal balls, and (d) multiplying by 6. Models frequently omit the legal delivery filter or use incorrect formulas, producing queries that execute successfully but return wrong values.

\paragraph{Entity Resolution Failures.}
Mapping player names to their correct \texttt{player\_id} values and associating players with specific teams poses a significant challenge. In international formats, this involves national team affiliations; in IPL, franchise affiliations that change across seasons. Models frequently fail to correctly join through the \texttt{PlayerInMatch} table when computing team-specific statistics.

\paragraph{Temporal Logic Errors.}
Queries involving temporal constraints (e.g., ``debut match,'' ``last 5 matches'') require careful use of \texttt{MIN(start\_date)} or window functions. Models frequently substitute \texttt{match\_id} ordering for date ordering or fail to apply correct \texttt{LIMIT} clauses for ``top-$N$'' queries.

\section{Conclusion}

This study introduces \textbf{CricBench}, a benchmark for evaluating intrinsic Text-to-SQL capability in cricket analytics using schema-only prompting across Test, ODI, T20I, and IPL. It reveals a substantial domain gap: \textbf{no single model dominates across formats}. In CricBench-Full, GPT-5 Mini leads on Test (12.4\% DMA) and ODI (10.9\%), while Claude Sonnet 4 is strongest on IPL (23.8\% on a reduced subset). In CricBench-Sub, DeepSeek R1 leads Test (33.0\%), and Qwen 235B leads T20I (17.5\%) and IPL (28.7\%). Even the best results remain far from practical use, with near-perfect execution but low semantic correctness; all seven models score 0\% on hard ODI queries.

Performance is also \textbf{highly inconsistent} across difficulty levels and formats: for example, DeepSeek R1 reaches 5\% on medium Test queries but only 4.5\% on T20I, while Qwen 235B drops from 31.8\% on medium T20I to 2.1\% on hard T20I, despite scoring 30.5\% on hard IPL. The domain gap analysis (Section~\ref{sec:domain_gap}) further shows a \textbf{37–55 percentage-point accuracy drop} from BIRD to CricBench, with the strongest general-purpose models suffering the largest declines, confirming that general Text-to-SQL ability does not transfer cleanly to specialized domains.

Looking forward, CricBench establishes a challenging baseline that current models clearly cannot solve through scale or prompting alone. Future work should explore domain-specific fine-tuning, structured retrieval-augmented generation with domain rule injection, and reasoning-enhanced architectures to close the substantial gap between current performance and practical utility in specialized analytics.

\section{Limitations}
CricBench does not yet support real-time live match queries. The evaluation of Indic languages is limited to English, Hindi, Punjabi, and Telugu, excluding other major languages like Tamil and Bengali. Additionally, our evaluation uses schema-only prompting; the impact of few-shot examples, retrieval-augmented generation, or domain-specific fine-tuning on CricBench performance remains unexplored and represents an important direction for future work. The varying sample sizes across models, particularly for CricBench-Sub models (DeepSeek R1 and Qwen 235B), limit direct comparability of aggregate metrics; we report per-format results with sample sizes throughout to ensure transparency. The small number of hard ODI queries ($N$=20) limits the statistical power of that particular finding. Our BIRD comparison (Section~\ref{sec:domain_gap}) uses the closest available model versions for GPT-5 Mini (GPT-4o), Claude Sonnet 4 (Claude 3.5/3.7 Sonnet), and Qwen 235B (Qwen 2.5 72B Instruct), which may slightly overstate or understate the true domain gap for those models.

\section{Data Availability}
To facilitate future research in domain-specific Text-to-SQL, we will make all artifacts from this study publicly available upon acceptance.

\paragraph{Released Assets}
The release package will include:
\begin{enumerate}[noitemsep]
    \item \textbf{CricBench Database:} The fully normalized SQLite databases for Test, ODI, T20I, and IPL formats containing granular ball-by-ball data.
    \item \textbf{Gold Standard Dataset:} The expert-verified dataset (\texttt{.json}) containing 633 sets of natural language queries (English, Hindi, Punjabi, \& Telugu) and their corresponding ground-truth SQL across all four formats.
    \item \textbf{Evaluation Harness:} The complete Python codebase for the execution engine and the metric calculation scripts (Execution Accuracy and Data Match Accuracy).
\end{enumerate}

\section*{Acknowledgments}
We acknowledge the use of Gemini (Google DeepMind) and Claude (Anthropic) for assistance in enhancing the grammatical precision and presentation of this paper. The experimental design, data curation, analysis, and all scientific conclusions accurately represent the original contributions of the authors.

\section*{Ethics Statement}
All data used in CricBench is derived from publicly available sports statistics. No personally identifiable information (PII) of players or fans is included beyond what is already in the public domain. The dataset is intended solely for research purposes to advance the field of specialized Text-to-SQL generation.

\bibliography{colm2026_conference}
\bibliographystyle{colm2026_conference}

\appendix
\section{Appendix: Prompting Specification}
\label{sec:appendix_prompts}

All models receive the identical system prompt shown below. This prompt provides the database schema and minimal instructions, with no calculation formulas or few-shot examples. Two lightweight domain hints (legal delivery definition and bowler wicket exclusions) are included to disambiguate schema semantics.

\begin{tcolorbox}[colback=gray!10, colframe=gray!40, title=Schema-Only System Prompt]
\small
\texttt{You are a SQL expert. Write a SQLite query using ONLY this schema:} \\[4pt]
\texttt{PRAGMA foreign\_keys=OFF;} \\[2pt]
\texttt{CREATE TABLE IF NOT EXISTS Players (} \\
\texttt{~~~~player\_id TEXT PRIMARY KEY,} \\
\texttt{~~~~player\_name TEXT NOT NULL} \\
\texttt{);} \\[2pt]
\texttt{CREATE TABLE IF NOT EXISTS Matches (} \\
\texttt{~~~~match\_id INTEGER PRIMARY KEY,} \\
\texttt{~~~~season TEXT NOT NULL,} \\
\texttt{~~~~start\_date TEXT NOT NULL,} \\
\texttt{~~~~end\_date TEXT,} \\
\texttt{~~~~venue TEXT NOT NULL,} \\
\texttt{~~~~city TEXT,} \\
\texttt{~~~~...} \textit{[full schema omitted for brevity]} \\
\texttt{);} \\[2pt]
\texttt{CREATE TABLE IF NOT EXISTS Deliveries (...);} \\
\texttt{CREATE TABLE IF NOT EXISTS PlayerInMatch (...);} \\
\texttt{CREATE TABLE IF NOT EXISTS FielderDismissals (...);} \\[4pt]
\texttt{PRAGMA foreign\_keys=ON;} \\[4pt]
\texttt{IMPORTANT:} \\
\texttt{- Use ONLY the tables and columns listed above.} \\
\texttt{- There is NO Innings table.} \\
\texttt{- Legal deliveries: wides = 0 AND noballs = 0.} \\
\texttt{- Bowler wickets exclude: `run out', `retired hurt',} \\
\texttt{~~`obstructing the field'.} \\
\texttt{- Output ONLY raw SQL (no explanation, no markdown).}
\end{tcolorbox}

\section{Appendix: Per-Format Detailed Results}
\label{sec:appendix_format}

Tables~\ref{tab:test_full}--\ref{tab:ipl_full} present detailed results broken down by language and difficulty for each format.

\begin{table}[H]
\begin{center}
\caption{Test Cricket -- Detailed Results (EX\% / DMA\%)}
\label{tab:test_full}
\scriptsize
\setlength{\tabcolsep}{3pt}
\begin{tabular}{l|cc|cc|cc|cc|cc|cc|cc}
\toprule
& \multicolumn{8}{c|}{\textbf{By Language}} & \multicolumn{6}{c}{\textbf{By Difficulty}} \\
\textbf{Model}
  & \multicolumn{2}{c|}{\textbf{EN}}
  & \multicolumn{2}{c|}{\textbf{HI}}
  & \multicolumn{2}{c|}{\textbf{PA}}
  & \multicolumn{2}{c|}{\textbf{TE}}
  & \multicolumn{2}{c|}{\textbf{Easy}}
  & \multicolumn{2}{c|}{\textbf{Med.}}
  & \multicolumn{2}{c}{\textbf{Hard}} \\
& EX & DMA & EX & DMA & EX & DMA & EX & DMA & EX & DMA & EX & DMA & EX & DMA \\
\midrule
GPT-5 Mini       & 99.4 & 13.0 & 99.4 & 14.8 & 100 & 11.2 & 100 & 10.7 & 100 & 13.6 & 99.7 & 11.4 & 99.5 & 13.0 \\
DeepSeek V3      & 99.4 & 14.2 & 100 & 13.0 & 100 & 11.8 & 100 & 8.3 & 100 & 11.9 & 100 & 11.1 & 99.5 & 13.0 \\
DeepSeek R1      & 96.0 & 32.0 & 100 & 24.0 & 100 & 36.0 & 96.0 & 40.0 & 100 & 25.0 & 93.8 & 50.0 & 100 & 25.0 \\
Qwen 235B         & 97.6 & 7.1  & 100 & 11.9 & 97.6 & 11.9 & 100 & 9.5 & 100 & 10.7 & 99.0 & 9.4 & 97.7 & 11.4 \\
Claude Son.\ 4   & 92.9 & 8.3  & 91.7 & 9.5  & 94.1 & 8.9 & 91.1 & 8.9 & 98.9 & 6.2 & 91.8 & 12.7 & 87.5 & 4.9 \\
Llama 3.1        & 95.3 & 5.3  & 95.9 & 4.1  & 94.7 & 4.1 & 95.3 & 1.2 & 98.9 & 12.5 & 92.1 & 0.9 & 97.3 & 0.0 \\
Gemma 2          & 57.4 & 3.6  & 62.7 & 1.8  & 66.3 & 1.8 & 84.6 & 2.4 & 88.6 & 5.1 & 63.3 & 1.9 & 55.4 & 0.5 \\
\bottomrule
\end{tabular}
\end{center}
\end{table}

\begin{table}[H]
\begin{center}
\caption{ODI Cricket -- Detailed Results (EX\% / DMA\%)}
\label{tab:odi_full}
\scriptsize
\setlength{\tabcolsep}{3pt}
\begin{tabular}{l|cc|cc|cc|cc|cc|cc|cc}
\toprule
& \multicolumn{8}{c|}{\textbf{By Language}} & \multicolumn{6}{c}{\textbf{By Difficulty}} \\
\textbf{Model}
  & \multicolumn{2}{c|}{\textbf{EN}}
  & \multicolumn{2}{c|}{\textbf{HI}}
  & \multicolumn{2}{c|}{\textbf{PA}}
  & \multicolumn{2}{c|}{\textbf{TE}}
  & \multicolumn{2}{c|}{\textbf{Easy}}
  & \multicolumn{2}{c|}{\textbf{Med.}}
  & \multicolumn{2}{c}{\textbf{Hard}} \\
& EX & DMA & EX & DMA & EX & DMA & EX & DMA & EX & DMA & EX & DMA & EX & DMA \\
\midrule
GPT-5 Mini       & 98.4 & 14.1 & 98.4 & 6.2  & 98.4 & 12.5 & 100 & 10.9 & 97.9 & 13.6 & 100 & 9.4 & 100 & 0.0 \\
DeepSeek V3      & 98.4 & 7.8  & 92.2 & 10.9 & 96.9 & 12.5 & 95.3 & 10.9 & 95.0 & 15.7 & 95.8 & 5.2 & 100 & 0.0 \\
DeepSeek R1      & 97.5 & 20.0 & 97.5 & 10.0 & 95.0 & 10.0 & 92.5 & 7.5 & 96.4 & 17.9 & 97.6 & 10.7 & 85.0 & 0.0 \\
Qwen 235B         & 27.5 & 7.5  & 27.5 & 2.5  & 27.5 & 7.5 & 27.5 & 5.0 & 50.0 & 14.3 & 19.0 & 1.2 & 0.0 & 0.0 \\
Claude Son.\ 4   & 95.3 & 12.5 & 93.8 & 1.6  & 92.2 & 6.2 & 93.8 & 4.7 & 91.4 & 6.4 & 96.9 & 7.3 & 95.0 & 0.0 \\
Llama 3.1        & 33.3 & 2.8  & 44.4 & 0.0  & 33.3 & 0.0 & 47.2 & 2.8 & 39.6 & 4.2 & 38.8 & 0.0 & 43.8 & 0.0 \\
Gemma 2          & 48.4 & 4.7  & 48.4 & 4.7  & 56.2 & 0.0 & 59.4 & 4.7 & 57.9 & 5.0 & 51.0 & 2.1 & 30.0 & 0.0 \\
\bottomrule
\end{tabular}
\end{center}
\end{table}

\begin{table}[H]
\begin{center}
\caption{T20 International -- Detailed Results (EX\% / DMA\%). GPT-5 Mini language and difficulty values are from a 399-instance subset.}
\label{tab:t20_full}
\scriptsize
\setlength{\tabcolsep}{3pt}
\begin{tabular}{l|cc|cc|cc|cc|cc|cc|cc}
\toprule
& \multicolumn{8}{c|}{\textbf{By Language}} & \multicolumn{6}{c}{\textbf{By Difficulty}} \\
\textbf{Model}
  & \multicolumn{2}{c|}{\textbf{EN}}
  & \multicolumn{2}{c|}{\textbf{HI}}
  & \multicolumn{2}{c|}{\textbf{PA}}
  & \multicolumn{2}{c|}{\textbf{TE}}
  & \multicolumn{2}{c|}{\textbf{Easy}}
  & \multicolumn{2}{c|}{\textbf{Med.}}
  & \multicolumn{2}{c}{\textbf{Hard}} \\
& EX & DMA & EX & DMA & EX & DMA & EX & DMA & EX & DMA & EX & DMA & EX & DMA \\
\midrule
GPT-5 Mini       & 98.0 & 5.0 & 100 & 3.0 & 100 & 3.0 & 99.0 & 5.1 & 100 & 0.0 & 100 & 8.6 & 98.6 & 1.4 \\
DeepSeek V3      & 84.0 & 9.5 & 85.5 & 9.5 & 87.0 & 7.0 & 86.4 & 5.5 & 87.5 & 1.0 & 95.4 & 10.3 & 74.6 & 7.2 \\
DeepSeek R1      & 94.0 & 2.0 & 96.0 & 6.0 & 98.0 & 6.0 & 98.0 & 4.1 & 96.9 & 3.1 & 98.5 & 5.9 & 94.0 & 4.5 \\
Qwen 235B         & 82.0 & 16.0 & 84.0 & 20.0 & 84.0 & 20.0 & 68.0 & 14.0 & 100 & 31.2 & 86.4 & 31.8 & 69.8 & 2.1 \\
Claude Son.\ 4   & 92.0 & 11.5 & 92.5 & 8.5 & 94.0 & 8.0 & 92.5 & 5.5 & 96.9 & 6.3 & 91.6 & 12.2 & 92.8 & 4.8 \\
Llama 3.1        & 14.0 & 0.5 & 19.5 & 0.5 & 22.0 & 0.5 & 21.1 & 0.0 & 20.8 & 0.0 & 22.0 & 0.3 & 15.5 & 0.6 \\
Gemma 2          & 54.0 & 1.5 & 56.0 & 0.5 & 59.5 & 0.5 & 54.8 & 1.0 & 69.8 & 0.0 & 59.5 & 0.8 & 48.4 & 1.2 \\
\bottomrule
\end{tabular}
\end{center}
\end{table}

\begin{table}[H]
\begin{center}
\caption{IPL -- Detailed Results (EX\% / DMA\%). Language values aggregate all Hindi and Telugu translation variants.}
\label{tab:ipl_full}
\scriptsize
\setlength{\tabcolsep}{3pt}
\begin{tabular}{l|cc|cc|cc|cc|cc|cc|cc}
\toprule
& \multicolumn{8}{c|}{\textbf{By Language}} & \multicolumn{6}{c}{\textbf{By Difficulty}} \\
\textbf{Model}
  & \multicolumn{2}{c|}{\textbf{EN}}
  & \multicolumn{2}{c|}{\textbf{HI}}
  & \multicolumn{2}{c|}{\textbf{PA}}
  & \multicolumn{2}{c|}{\textbf{TE}}
  & \multicolumn{2}{c|}{\textbf{Easy}}
  & \multicolumn{2}{c|}{\textbf{Med.}}
  & \multicolumn{2}{c}{\textbf{Hard}} \\
& EX & DMA & EX & DMA & EX & DMA & EX & DMA & EX & DMA & EX & DMA & EX & DMA \\
\midrule
GPT-5 Mini       & 99.5 & 14.0 & 97.4 & 9.6  & 96.5 & 11.5 & 96.8 & 9.5  & 98.4 & 16.7 & 97.8 & 8.9  & 96.4 & 10.3 \\
DeepSeek V3      & 95.5 & 9.5  & 94.0 & 9.6  & 96.0 & 6.0  & 95.9 & 6.4  & 99.0 & 13.5 & 96.2 & 7.3  & 91.1 & 5.3 \\
DeepSeek R1      & 96.0 & 16.0 & 94.7 & 12.0 & 98.0 & 12.0 & 94.5 & 10.9 & 100 & 34.4 & 95.5 & 2.7  & 91.5 & 8.5 \\
Qwen 235B         & 84.0 & 30.0 & 96.0 & 33.3 & 96.0 & 28.0 & 76.4 & 21.8 & 93.4 & 29.5 & 94.5 & 27.3 & 72.9 & 30.5 \\
Claude Son.\ 4   & 80.0 & 23.3 & 82.5 & 22.5 & 90.0 & 23.3 & 80.0 & 26.7 & 87.8 & 31.7 & 90.9 & 27.3 & 71.1 & 13.3 \\
Llama 3.1        & 50.5 & 0.5  & 60.9 & 0.3  & 56.0 & 1.0  & 55.0 & 0.5  & 59.9 & 2.6  & 56.8 & 0.0  & 52.7 & 0.0 \\
Gemma 2          & 23.5 & 0.0  & 32.5 & 0.0  & 38.0 & 0.5  & 43.2 & 0.9  & 38.0 & 1.6  & 36.7 & 0.0  & 27.8 & 0.0 \\
\bottomrule
\end{tabular}
\end{center}
\end{table}

\section{Appendix: Example Queries}
\label{sec:appendix_examples}

Tables~\ref{tab:examples_test}--\ref{tab:examples_ipl} provide illustrative examples of CricBench queries at each difficulty level with their corresponding SQL. These examples demonstrate the range of domain knowledge and SQL complexity required.

\begin{table}[H]
\begin{center}
\caption{Example Test Cricket Queries}
\label{tab:examples_test}
\small
\begin{tabular}{p{1.2cm}p{5.5cm}p{6cm}}
\toprule
\textbf{Diff.} & \textbf{Query} & \textbf{SQL (abbreviated)} \\
\midrule
Easy & \textbf{EN:} Who scored the most runs in Test matches at Lord's? \newline \textbf{HI:} Lord's mein Test matches mein sabse zyada runs kisne banaye? & \texttt{SELECT p.player\_name, SUM(d.runs\_batter) AS total FROM Deliveries d JOIN ... WHERE m.venue = "Lord's" GROUP BY ... ORDER BY total DESC LIMIT 1} \\
\midrule
Medium & \textbf{EN:} Which bowler has the best economy rate in Test matches since 2015, with at least 500 legal deliveries? & \texttt{SELECT p.player\_name, SUM(d.runs\_total)*6.0 / COUNT(*) AS econ FROM Deliveries d JOIN ... WHERE d.wides=0 AND d.noballs=0 AND m.start\_date >= '2015' GROUP BY ... HAVING COUNT(*) >= 500 ORDER BY econ LIMIT 1} \\
\midrule
Hard & \textbf{EN:} For each team, find the player with the highest strike rate in the first innings of matches they won, min 100 balls faced. & \texttt{WITH cte AS (SELECT ... SUM(runs)*100.0 / COUNT(*) AS sr ... WHERE d.wides=0 AND d.noballs=0 AND d.innings=1 ... HAVING COUNT(*)>=100) SELECT ... RANK() OVER ...} \\
\bottomrule
\end{tabular}
\end{center}
\end{table}

\begin{table}[H]
\begin{center}
\caption{Example T20I Queries}
\label{tab:examples_t20}
\small
\begin{tabular}{p{1.2cm}p{5.5cm}p{6cm}}
\toprule
\textbf{Diff.} & \textbf{Query} & \textbf{SQL (abbreviated)} \\
\midrule
Easy & \textbf{EN:} How many T20I matches has India played? \newline \textbf{PA:} India ne kitne T20I matches khele han? & \texttt{SELECT COUNT(DISTINCT m.match\_id) FROM Matches m JOIN PlayerInMatch pim ON ... WHERE pim.team = 'India'} \\
\midrule
Medium & \textbf{EN:} Top 5 batters by runs in powerplay overs (1--6) in T20I matches in 2023. & \texttt{SELECT p.player\_name, SUM(d.runs\_batter) AS pp\_runs FROM ... WHERE d.over\_number BETWEEN 1 AND 6 AND m.season = '2023' GROUP BY ... ORDER BY pp\_runs DESC LIMIT 5} \\
\midrule
Hard & \textbf{EN:} Find players who have hit at least 3 centuries and also taken at least 10 wickets in T20I. & \texttt{WITH batting AS (...HAVING SUM(runs\_batter)>=100...), bowling AS (...WHERE d.is\_wicket=1 AND d.dismissal\_kind NOT IN ('run out',...) HAVING COUNT(*)>=10) SELECT ... FROM batting JOIN bowling ...} \\
\bottomrule
\end{tabular}
\end{center}
\end{table}

\begin{table}[H]
\begin{center}
\caption{Example IPL Queries}
\label{tab:examples_ipl}
\small
\begin{tabular}{p{1.2cm}p{5.5cm}p{6cm}}
\toprule
\textbf{Diff.} & \textbf{Query} & \textbf{SQL (abbreviated)} \\
\midrule
Easy & \textbf{EN:} Which team has won the most IPL matches? \newline \textbf{TE:} Ekkuva IPL matches gelicchina team edi? & \texttt{SELECT winner, COUNT(*) AS wins FROM Matches WHERE winner IS NOT NULL GROUP BY winner ORDER BY wins DESC LIMIT 1} \\
\midrule
Medium & \textbf{EN:} List the top 5 players by number of sixes hit across all IPL seasons. & \texttt{SELECT p.player\_name, COUNT(*) AS sixes FROM Deliveries d JOIN ... WHERE d.runs\_batter = 6 GROUP BY ... ORDER BY sixes DESC LIMIT 5} \\
\midrule
Hard & \textbf{EN:} For each season, find the player with the best bowling strike rate (min 20 wickets in that season). & \texttt{WITH season\_bowling AS (... COUNT(CASE WHEN is\_wicket=1 ...) AS wkts, COUNT(*) AS balls ... HAVING wkts>=20) SELECT season, player\_name, balls*1.0/wkts AS bowl\_sr ... RANK() OVER(PARTITION BY season ORDER BY bowl\_sr) ...} \\
\bottomrule
\end{tabular}
\end{center}
\end{table}

\begin{table}[H]
\begin{center}
\caption{Multilingual Query Examples (same query across languages)}
\label{tab:examples_multilingual}
\small
\begin{tabular}{lp{10cm}}
\toprule
\textbf{Language} & \textbf{Query} \\
\midrule
English & Which player has the highest batting average in ODI matches played in India? \\
Hindi & India mein khele gaye ODI matches mein sabse zyada batting average kiska hai? \\
Punjabi & India vich khede gaye ODI matches vich sabto vadh batting average kihda hai? \\
Telugu & India lo aadina ODI matches lo andari kanna ekkuva batting average evari di? \\
\bottomrule
\end{tabular}
\end{center}
\end{table}

\end{document}